\documentclass[11pt]{article}

\pdfoutput=1 

\usepackage{naaclhlt2016}
\usepackage{times}
\usepackage{url}
\usepackage{latexsym}
\usepackage{flushend} 

\naaclfinalcopy 

\AddToShipoutPicture{\AtPageLowishCenter{\thepage}}

\usepackage{amsfonts,eucal,amsbsy,amsopn}
\usepackage{amsmath,amssymb,amsthm,stmaryrd,color}
\usepackage{graphicx}
\usepackage{multirow,rotating}
\usepackage{xspace}
\usepackage{dsfont}
\usepackage{url}
\usepackage{verbatim}
\usepackage[font=small]{caption}
\usepackage{mdframed}

\makeatletter
\newcommand{\@BIBLABEL}{\@emptybiblabel}
\newcommand{\@emptybiblabel}[1]{}
\makeatother

\usepackage[hidelinks,pdfusetitle]{hyperref}

\makeatletter
\newcommand*{\hyperlinkcite}[1]{\hyper@link{cite}{cite.#1}}
\makeatother

\usepackage[tight]{subfigure}
\usepackage{subfloat}
\usepackage{wrapfig}

\usepackage{tabularx}
\usepackage{booktabs}

\newcommand{\finalversion}[1]{}

\newtheorem{lemma-ap}{Lemma}

\newcommand{\ignore}[1]{}
\newcommand{\argmax}[2]{\underset{#1}{\textrm{arg max }}\xspace #2}

\def\figref#1{Fig.~\ref{#1}}
\def\secref#1{Sec.~\ref{#1}}

\def\eqnref#1{Eqn.~\ref{#1}}

\begin{document}
\title{What to talk about and how? Selective Generation using\\ LSTMs with Coarse-to-Fine Alignment}

\author{
Hongyuan Mei \qquad Mohit Bansal \qquad Matthew R.~Walter\\
Toyota Technological Institute at Chicago\\
Chicago, IL 60637 \\
\{\texttt{hongyuan,mbansal,mwalter\}@ttic.edu}
}

\maketitle

\begin{abstract}
    We propose an end-to-end, domain-independent neural
    encoder-aligner-decoder model for selective generation, i.e., the  
    joint task of content selection and surface realization. Our model
    first encodes a full set of over-determined database event
    records via an
    LSTM-based recurrent neural network, then utilizes a novel
    coarse-to-fine aligner to identify the small
    subset of salient records to talk about, and finally employs a
    decoder to generate free-form descriptions of the aligned, selected
    records. Our model achieves the best selection and generation
    results reported to-date (with  $59\%$ relative improvement in
    generation) on the benchmark \textsc{WeatherGov}
    dataset, despite using no specialized features or
    linguistic resources. 
    Using an improved $k$-nearest neighbor beam filter helps further.
We also perform a series of ablations and visualizations to elucidate the
contributions of our key model components. Lastly, we evaluate the generalizability of our model on the
    \textsc{RoboCup} dataset, and get results that are
    competitive with 
    or better than the state-of-the-art, despite being severely
    data-starved.
\end{abstract}

\section{Introduction} \label{sec:introduction}

We consider the important task of producing a natural language
description of a rich world state represented as an over-determined
database of event records. This task, which we refer to as selective
generation, is often formulated as two subproblems: \emph{content
  selection}, which involves choosing a subset of relevant records to
talk about from the exhaustive database, and \emph{surface
  realization}, which is concerned with generating natural language
descriptions for this subset. Learning to perform these tasks jointly
is challenging due to the ambiguity in deciding which records are
relevant, the complex dependencies between selected records, and the
multiple ways in which these records can be described.

Previous work has made significant progress on this
task~\cite{chen-08,angeli-10,kim-10,konstas-12}.  However, most
approaches solve the two content selection and surface realization
subtasks separately, use manual domain-dependent resources (e.g.,
semantic parsers) and features, or employ template-based
generation. This limits domain adaptability and reduces coherence. We
take an alternative, neural encoder-aligner-decoder approach to
free-form selective generation that jointly performs content selection
and surface realization, without using any specialized features,
resources, or generation templates. This enables our approach to
generalize to new domains.  Further, our memory-based model captures
the long-range contextual dependencies among records and descriptions,
which are integral to this task~\cite{angeli-10}.

We formulate our model as an encoder-aligner-decoder framework that
uses recurrent neural networks with long short-term memory units
(LSTM-RNNs)~\cite{hochreiter-97} together with a coarse-to-fine
aligner to select and ``translate'' the rich world state into a
natural language description. Our model first encodes the full set of
over-determined event records using a bidirectional LSTM-RNN. A novel
coarse-to-fine aligner then reasons over multiple abstractions of the
input to decide which of the records to discuss. The model next
employs an LSTM decoder to generate natural language descriptions of
the selected records.

The use of LSTMs, which have proven effective for similar long-range
generation tasks~\cite{sutskever-14,vinyals-14,karpathy-15}, allows
our model to capture the long-range contextual dependencies that
exist in selective generation. Further, the introduction of
our proposed variation on alignment-based
LSTMs~\cite{bahdanau-14,xu-15} enables our model to learn to perform
content selection and surface realization jointly, by aligning each
generated word to an event record during decoding.  Our novel
coarse-to-fine aligner avoids searching over the full set of
over-determined records by employing two stages of increasing
complexity: a pre-selector and a refiner acting on multiple
abstractions (low- and high-level) of the record input.  The
end-to-end nature of our framework has the advantage that it can be
trained directly on corpora of record sets paired with natural
language descriptions, without the need for ground-truth content
selection.

We evaluate our model on a benchmark weather forecasting dataset
(\textsc{WeatherGov}) and achieve the best results reported to-date on
content selection ($12\%$ relative improvement in F-1) and language
generation ($59\%$ relative improvement in BLEU), despite using no
domain-specific resources. We also perform a series of ablations and
visualizations to elucidate the contributions of the primary model
components, and also show improvements with a simple, $k$-nearest
neighbor beam filter approach. Finally, we demonstrate the
generalizability of our model by directly applying it to a benchmark
sportscasting dataset (\textsc{RoboCup}), where we get results
competitive with or better than state-of-the-art, despite being
extremely data-starved.
\section{Related Work} \label{sec:related}

Selective generation is a relatively new research area and more
attention has been paid to the individual content selection and
selective realization subproblems. With regards to the former,
\newcite{barzilay-04} model the content structure from unannotated
documents and apply it to the application of text
summarization. \newcite{barzilay-05} treat content selection as a
collective classification problem and simultaneously optimize the
local label assignment and their pairwise
relations. \newcite{liang-09} address the related task of aligning a
set of records to given textual description clauses. They propose a
generative semi-Markov alignment model that jointly segments text
sequences into utterances and associates each to the corresponding
record.

Surface realization is often treated as a problem of producing text
according to a given grammar.  \newcite{soricut-06} propose a language
generation system that uses the WIDL-representation, a formalism used
to compactly represent probability distributions over finite sets of
strings.  \newcite{wong-07} and \newcite{lu-11} use synchronous
context-free grammars to generate natural language sentences from
formal meaning representations. Similarly, \newcite{belz-08} employs
probabilistic context-free grammars to perform surface
realization. Other effective approaches include the use of tree
conditional random fields~\cite{lu-09} and template extraction within
a log-linear framework~\cite{angeli-10}.

Recent work seeks to solve the full selective generation problem
through a single framework.  \newcite{chen-08} and~\newcite{chen-10}
learn alignments between comments and their corresponding event
records using a translation model for parsing and
generation. \newcite{kim-10} implement a two-stage framework that
decides what to discuss using a combination of the methods of
\newcite{lu-08} and \newcite{liang-09}, and then produces the text
based on the generation system of \newcite{wong-07}.

\newcite{angeli-10} propose a unified concept-to-text model that
treats joint content selection and surface realization as a sequence
of local decisions represented by a log-linear model. Similar to other
work, they train their model using external alignments from
\newcite{liang-09}. Generation then follows as inference over this
model, where they first choose an event record, then the record's
fields (i.e., attributes), and finally a set of templates that they
then fill in with words for the selected fields. Their ability to
model long-range dependencies relies on their choice of features for
the log-linear model, while the template-based generation further
employs some domain-specific features for fluent output.

\newcite{konstas-12} propose an alternative method that simultaneously
optimizes the content selection and surface realization problems. They
employ a probabilistic context-free grammar that specifies the
structure of the event records, and then treat generation as finding
the best derivation tree according to this grammar. However, their
method still selects and orders records in a local fashion via a
Markovized chaining of records.  \newcite{konstas-13} improve upon
this approach with global document representations.  However, this
approach also requires alignment during training, which they estimate
using the method of~\newcite{liang-09}.

We treat the problem of selective generation as end-to-end learning
via a recurrent neural network encoder-aligner-decoder model, which
enables us to jointly learn content selection and surface realization
directly from database-text pairs, without the need for an external
aligner or ground-truth selection labels. The use of LSTM-RNNs enables
our model to capture the long-range dependencies that exist among the
records and natural language output. Additionally, the model does not
rely on any manually-selected or domain-dependent features, templates,
or parsers, and is thereby generalizable.
The alignment-RNN approach has recently proven successful for
generation-style tasks, e.g., machine translation~\cite{bahdanau-14}
and image captioning~\cite{xu-15}. Since selective generation requires
identifying the small number of salient records among an
over-determined database, we avoid performing exhaustive search over
the full record set, and instead propose a novel coarse-to-fine
aligner that divides the search complexity into pre-selection and
refinement stages.


\section{Task Definition} \label{sec:taskdefinition} We consider the
problem of generating a natural language description for a rich world
state specified in terms of an over-determined set of records
(database).  This problem requires deciding which of the records to
discuss (content selection) and how to discuss them (surface
realization).  Training data consists of \emph{scenario} pairs
$(r^{(i)}, x^{(i)})$ for $i = 1,2,\ldots,n$, where $r^{(i)}$ is the
complete set of records and $x^{(i)}$ is the natural language
description (\figref{fig:example_input}). At test time, only the
records are given.  We evaluate our model in the context of two
publicly-available benchmark selective generation datasets.
\begin{figure}[!tb]
    \centering
    \subfigure[\textsc{WeatherGov}]{%
      \begin{minipage}[t]{1.0\linewidth}
          \begin{minipage}[!c]{0.1\linewidth}
              $r_{1:N}$:
          \end{minipage}
          \begin{minipage}[!t]{0.9\linewidth}
              \begin{mdframed}[align=center]{\scriptsize
                    temperature(time=\texttt{17-06}, min=\texttt{48},
                    mean=\texttt{53}, max=\texttt{61})\\ 
                    windSpeed(time=\texttt{17-06}, min=\texttt{3},
                    mean=\texttt{6}, max=\texttt{11})\\
                    windDir(time=\texttt{17-06}, mode=\texttt{SSW}) \\
                    gust(time=\texttt{17-06}, min=\texttt{0},
                    mean=\texttt{0}, max=\texttt{0}) \\
                    skyCover(time=\texttt{17-21}, mode=\texttt{0-25}) \\
                    skyCover(time=\texttt{02-06}, mode=\texttt{75-100}) \\
                    precipChance(time=\texttt{17-06}, min=\texttt{2},
                    mean=\texttt{14}, max=\texttt{20}) \\
                    rainChance(time=\texttt{17-06}, mode=\texttt{someChance})}
              \end{mdframed}
          \end{minipage}
          \begin{minipage}[!c]{0.1\linewidth}
              $x_{1:N}$:
          \end{minipage}
          \begin{minipage}[!t]{0.9\linewidth}
              \begin{mdframed}[align=center,hidealllines=true]
              {\scriptsize
                ``a 20 percent chance of showers after
                midnight. increasing clouds, with a low around 48
                southwest wind between 5 and 10 mph''}
              \end{mdframed}
          \end{minipage}
      \end{minipage}\label{fig:weathergov_example}}
    \subfigure[\textsc{RoboCup}]{%
      \begin{minipage}[t]{1.0\linewidth}
          \begin{minipage}[!c]{0.1\linewidth}
              $r_{1:N}$:
          \end{minipage}
          \begin{minipage}[!t]{0.9\linewidth}
              \begin{mdframed}[align=center]{\scriptsize
                    pass(arg1=\texttt{purple6}, arg2=\texttt{purple3})\\
                    kick(arg1=\texttt{purple3})\\
                    badPass(arg1=\texttt{purple3}, arg2=\texttt{pink9})\\
                    turnover(arg1=\texttt{purple3}, arg2=\texttt{pink9})}
              \end{mdframed}
          \end{minipage}
          \begin{minipage}[!c]{0.1\linewidth}
              $x_{1:N}$:
          \end{minipage}
          \begin{minipage}[!t]{0.9\linewidth}
              \begin{mdframed}[align=center,hidealllines=true]
              {\scriptsize
                ``purple3 made a bad pass that was picked off by pink9''}
              \end{mdframed}
          \end{minipage}
      \end{minipage}\label{fig:robocup_example}}
    \caption{Sample database-text pairs chosen from the
      \subref{fig:weathergov_example}~\textsc{WeatherGov} and
      \subref{fig:robocup_example}~\textsc{RoboCup} datasets.} \label{fig:example_input}
\end{figure}
\begin{figure}[ht]
    \centering
    \includegraphics[width=0.975\linewidth]{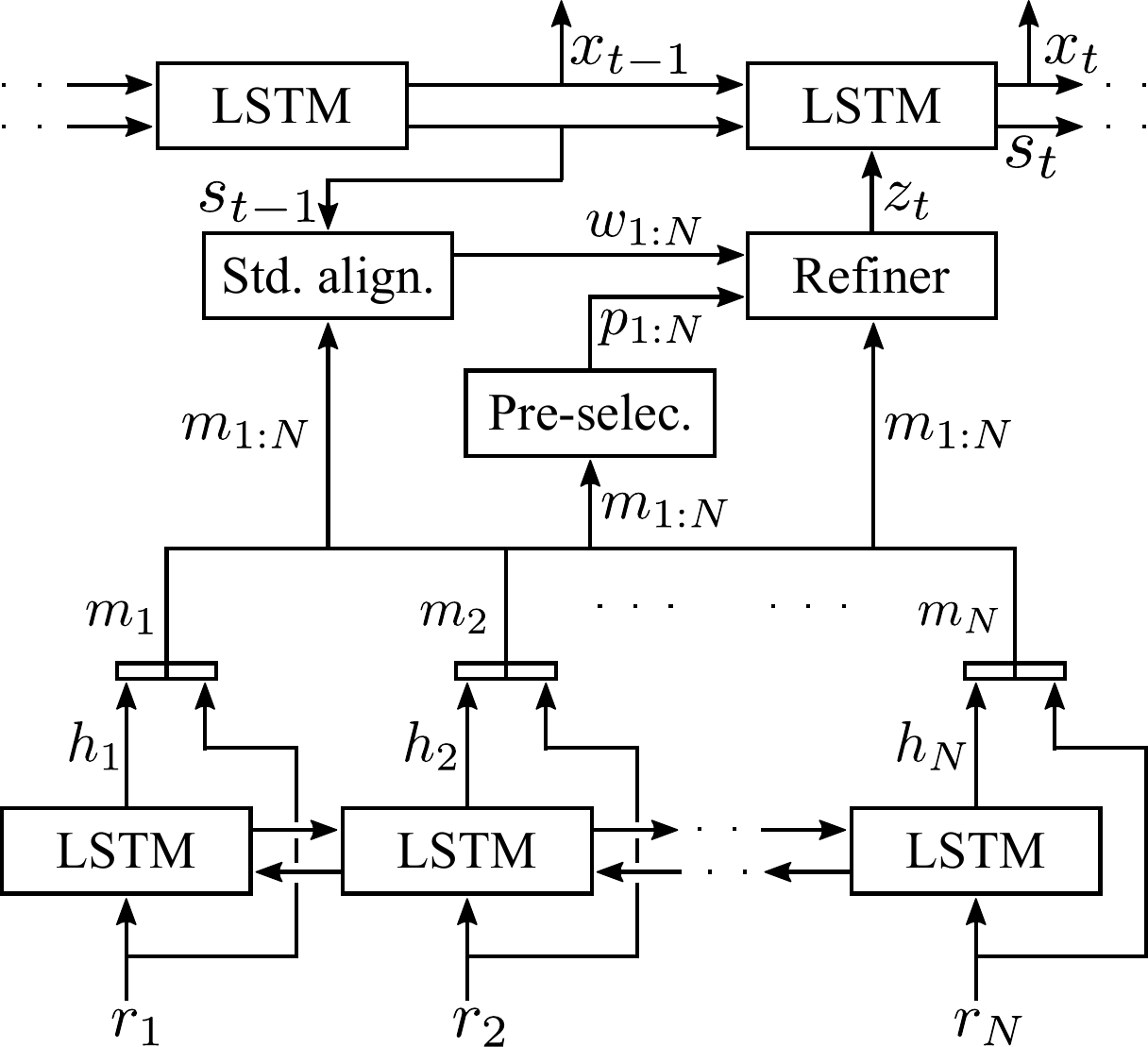}
    \caption{Our model architecture with a bidirectional LSTM encoder, coarse-to-fine aligner, and decoder.}
    \label{fig:modelfig}
\end{figure}

\paragraph{\textsc{WeatherGov}} 

The weather forecasting dataset (see
Fig.~\ref{fig:weathergov_example}) of~\newcite{liang-09} consists of
$29528$ scenarios, each with $36$ weather records (e.g., temperature,
sky cover, etc.)  paired with a natural language forecast ( $28.7$
avg. word length).

\paragraph{\textsc{RoboCup}}
We evaluate our model's generalizability on the sportscasting dataset
of~\newcite{chen-08}, which consists of only $1539$ pairs of
temporally ordered robot soccer events (e.g., pass, score) and
commentary drawn from the four-game 2001--2004 RoboCup finals (see
Fig.~\ref{fig:robocup_example}). Each scenario contains an average of
$2.4$ event records and a $5.7$ word natural language commentary.

\section{The Model}
\label{sec:model}

We formulate selective generation as inference over a probabilistic model
$P(x_{1:T} \vert r_{1:N})$, where
\mbox{$r_{1:N} = (r_1, r_2, \ldots, r_N)$} is the input set of
over-determined event records,\footnote{These records may take the form of
  an unordered set or have a natural ordering (e.g., temporal in the case
  of \textsc{RoboCup}). In order to make our model generalizable, we treat
  the set as a sequence and use the order specified by the dataset. We note
  that it is possible that a different ordering will yield improved
  performance, since ordering has been shown to be important when operating
  on sets~\cite{vinyals-15}.} $x_{1:T} = (x_1, x_2, \ldots, x_T)$ is the
generated description with $x_t$ being the word at time $t$ and $x_0$ being
a special start token:
\begin{subequations} \label{eqn:argmax}
	\begin{align}
    x_{1:T}^* &= \argmax{x_{1:T}}{P(x_{1:T} 
                                  \vert r_{1:N})}\\
    &= \underset{x_{1:T} }{\textrm{arg max }}
      \prod\limits_{t=1}^T P(x_t \vert x_{0:t-1}, r_{1:N})
	\end{align}
\end{subequations}

The goal of inference is to generate a natural language description for a
given set of records. An effective means of learning to perform this
generation is to use an encoder-aligner-decoder architecture with a
recurrent neural network, which has proven effective for related problems
in machine translation \cite{bahdanau-14} and image captioning
\cite{xu-15}. We propose a variation on this general model with novel
components that are well-suited to the selective generation problem.

Our model (\figref{fig:modelfig}) first encodes each input record $r_j$
into a hidden state $h_j$ with \mbox{$j \in \{1,\ldots,N\}$} using a
bidirectional recurrent neural network (RNN). Our novel coarse-to-fine
aligner then acts on a concatenation $m_j$ of each record and its hidden
state as multi-level representation of the input to compute the selection
decision $z_t$ at each decoding step $t$. The model then employs an RNN
decoder to arrive at the word likelihood
\mbox{$P(x_t \vert x_{0:t-1}, r_{1:N})$} as a function of the multi-level
input and the hidden state of the decoder $s_{t-1}$ at time step $t-1$. In
order to model the long-range dependencies among the records and
descriptions (which is integral to effectively performing selective
generation~\cite{angeli-10,konstas-12,konstas-13}), our model employs LSTM
units as the nonlinear encoder and decoder functions.

\paragraph{Encoder}
Our LSTM-RNN encoder (Fig.~\ref{fig:modelfig}) takes as input the set of
event records represented as a sequence
\mbox{$r_{1:N} = (r_1, r_2, \ldots, r_N)$} and returns a sequence of hidden
annotations \mbox{$h_{1:N} = (h_1,h_2,\ldots,h_N)$}, where the annotation
$h_j$ summarizes the record $r_j$. This results in a representation that
models the dependencies that exist among the records in the
database.

We adopt an encoder architecture similar to that
of~\newcite{graves-13}
\begin{subequations} \label{eqn:encoder}
    \begin{align}
      \begin{pmatrix}
          i^{e}_j\\ 
          f^{e}_j\\
          o^{e}_j\\ 
          g^{e}_j 
      \end{pmatrix} 
   &= 
     \begin{pmatrix}
         \sigma\\
         \sigma\\
         \sigma\\ 
         \tanh 
     \end{pmatrix}
      T^{e} 
      \begin{pmatrix}
          r_j\\ 
          h_{j-1}
      \end{pmatrix}\\
      c^{e}_j&=f^{e}_j \odot c^{e}_{j-1} + i^{e}_j \odot g^{e}_j \\
      h_j &=o^{e}_j \odot \tanh(c^{e}_j) \label{eqn:encoder_h}
    \end{align}
\end{subequations}
where $T^{e}$ is an affine transformation, $\sigma$ is the logistic sigmoid
that restricts its input to $[0,1]$, $i^{e}_j$, $f^{e}_j$, and $o^{e}_j$
are the input, forget, and output gates of the LSTM, respectively, and
$c^e_j$ is the memory cell activation vector. The memory cell $c^e_j$
summarizes the LSTM's previous memory $c^e_{j-1}$ and the current input,
which are modulated by the forget and input gates, respectively. Our
encoder operates bidirectionally, encoding the records in both the forward
and backward directions, which provides a better summary of the input
records. In this way, the hidden annotations
$h_j = (\overrightarrow{h}_j^\top; \overleftarrow{h}_j^\top)^\top$
concatenate forward $\overrightarrow{h}_j$ and backward
$\overleftarrow{h}_j$ annotations, each determined using
Equation~\eqref{eqn:encoder_h}.

\paragraph{Coarse-to-Fine Aligner}
Having encoded the input records $r_{1:N}$ to arrive at the hidden
annotations $h_{1:N}$, the model then seeks to select the content at each
time step $t$ that will be used for generation. Our model performs content
selection using an extension of the alignment mechanism proposed by
\newcite{bahdanau-14}, which allows for selection and generation that is
independent of the ordering of the input.

In selective generation, the given set of event records is over-determined
with only a small subset of salient records being relevant to the output
natural language description. Standard alignment mechanisms limit the
accuracy of selection and generation by scanning the entire range of
over-determined records.  In order to better address the selective
generation task, we propose a coarse-to-fine aligner that prevents the
model from being distracted by non-salient records. Our model aligns based
on multiple abstractions of the input: both the original input record as
well as the hidden annotations \mbox{$m_j = (r_j^\top; h_j^\top)^\top$}, an
approach that has previously been shown to yield better results than
aligning based only on the hidden state~\cite{mei-15}.

Our coarse-to-fine aligner avoids searching over the full set of
over-determined records by using two stages of increasing
complexity: a pre-selector and refiner (\figref{fig:modelfig}). The
pre-selector first assigns to each record a probability $p_j$ of being
selected, while the standard aligner computes the
alignment likelihood $w_{tj}$ over all the records at each time step $t$
during decoding. Next, the refiner produces the final selection decision by
re-weighting the aligner weights $w_{tj}$ with the pre-selector
probabilities $p_j$:
\begin{subequations} \label{eqn:aligner}
	\begin{align}
    p_{j} &= \text{sigmoid}\left( q^\top\tanh(P m_j) \right)\\
    \beta_{tj} &= v^\top\text{tanh} (Ws_{t-1}+Um_{j})\\
    w_{tj} &= \exp(\beta_{tj})/\sum_j \exp(\beta_{tj})\\
    \alpha_{tj} &= p_j w_{tj} /\sum_j p_j w_{tj}\\
    z_t &= \sum_{j}\alpha_{tj} m_j \label{eqn:wtdavg}
	\end{align}
\end{subequations}
where $P$, $q$, $U$, $W$, $v$ are learned parameters.
Ideally, the selection decision would be based on the highest-value
alignment $z_t = m_k$ where \mbox{$k = \textrm{arg max}_j \, \alpha_{tj}$}.
However, we use the weighted average (\eqnref{eqn:wtdavg}) as its soft
approximation to maintain differentiability of the entire architecture.

The pre-selector assigns large values ($p_j > 0.5$) to a small subset of
salient records and small values ($p_j < 0.5$) to the rest. This modulates
the standard aligner, which then has to assign a large weight $w_{tj}$ in
order to select the $j$-th record at time $t$. In this way, the learned
prior $p_j$ makes it difficult for the alignment (attention) to be
distracted by non-salient records. Further, we can relate the output of the
pre-selector to the number of records that are selected. Specifically, the
output $p_j$ expresses the extent to which the $j$-th record should be
selected. The summation $\sum_{j=1}^N p_j$ can then be regarded as a
real-valued approximation to the total number of pre-selected records
(denoted as $\gamma$), which we regularize towards, based on validation
(see Eqn.~\ref{eqn:lossreg}).

\paragraph{Decoder} 
Our architecture uses an LSTM decoder that takes as input the current
context vector $z_t$, the last word $x_{t-1}$, and the LSTM's previous
hidden state $s_{t-1}$.
The decoder outputs the conditional probability distribution $P_{x,t} = P(x_t \vert x_{0:t-1}, r_{1:N})$ over the next word, represented as a deep output layer~\cite{pascanu-14},
\begin{subequations}
\label{eqn:decoder}
    \begin{align}
      \begin{pmatrix}
          i^d_t\\
          f^d_t\\
          o^d_t\\
          g^d_t
      \end{pmatrix}
   &= 
     \begin{pmatrix}
         \sigma\\ 
         \sigma\\ 
         \sigma\\ 
         \tanh 
     \end{pmatrix} T^d 
      \begin{pmatrix}
          Ex_{t-1}\\ 
          s_{t-1}\\ 
          z_t
      \end{pmatrix}\\
      c^d_t&=f^d_t \odot c^d_{t-1} + i^d_t \odot g^d_t \\
      s_{t}&=o_t^d \odot \tanh(c^d_t)\\
      l_t &= L_0 ( Ex_{t-1} + L_ss_t + L_zz_t )\\
      P_{x,t} &= \textrm{softmax}\left(l_t\right)
    \end{align}
\end{subequations}
where $E$ (an embedding matrix), $L_0$, $L_s$, and $L_z$ are parameters to
be learned.

\paragraph{Training and Inference} 
We train the model using the database-record pairs $(r_{1:N}, x_{1:T})$
from the training corpora so as to maximize the likelihood of the
ground-truth language description $x^*_{1:T}$
(Eqn.~\ref{eqn:argmax}). Additionally, we introduce a regularization term
$(\sum_{j=1}^N p_j - \gamma)^2$ that enables the model to influence the
pre-selector weights based on the aforementioned relationship between the
output of the pre-selector and the number of selected records.  Moreover,
we also introduce the term $(1.0 - \max (p_j))$, which accounts for the
fact that at least one record should be pre-selected.  Note that when
$\gamma$ is equal to $N$, the pre-selector is forced to select all the
records ($p_j = 1.0$ for all $j$), and the coarse-to-fine alignment reverts
to the standard alignment introduced by~\newcite{bahdanau-14}. Together
with the negative log-likelihood of the ground-truth description
$x^*_{1:T}$, our loss function becomes
\begin{subequations}\label{eqn:lossreg}
    \begin{align}
      L &= -\log P(x^*_{1:T} \vert r_{1:N}) + G\\
        &= -\sum\limits_{t=1}^T \log P(x^*_t \vert x_{0:t-1}, r_{1:N}) +
          G\\
      G &= \left(\sum_{j=1}^N p_j - \gamma\right)^2 + \Bigl(1-\max (p_j)\Bigr) 
    \end{align} 
\end{subequations}

Having trained the model, we generate the natural language description by
finding the maximum a posteriori words under the learned model
(\eqnref{eqn:argmax}). For inference, we perform greedy search starting
with the first word $x_1$. Beam search offers a way to perform approximate
joint inference --- however, we empirically found that beam search does not
perform any better than greedy search on the datasets that we consider, an
observation that is shared with previous work~\cite{angeli-10}. We later
discuss an alternative $k$-nearest neighbor-based beam filter (see
Sec~\ref{sec:extensions}).

\section{Experimental Setup} \label{sec:experimentalsetup}

\paragraph{Datasets}
We analyze our model on the benchmark \textsc{WeatherGov} dataset, and
use the data-starved \textsc{RoboCup} dataset to demonstrate the
model's generalizability. Following \newcite{angeli-10}, we use
\textsc{WeatherGov} training, development, and test splits of size
$25000$, $1000$, and $3528$, respectively. For \textsc{RoboCup}, we
follow the evaluation methodology of previous work~\cite{chen-08},
performing three-fold cross-validation whereby we train on three games
(approximately $1000$ scenarios) and test on the fourth. Within each
split, we hold out $10\%$ of the training data as the development set
to tune the early-stopping criterion and $\gamma$. We then report the
standard average performance (weighted by the number of scenarios)
over these four splits.

\paragraph{Training Details}
On \textsc{WeatherGov}, we lightly tune the number of hidden units and
$\gamma$ on the development set according to the generation metric
(BLEU), and choose $500$ units from $\{250, 500, 750\}$ and
$\gamma = 8.5$ from $\{6.5, 7.5, 8.5, 10.5, 12.5\}$. For
\textsc{RoboCup}, we only tune $\gamma$ on the development set and
choose $\gamma = 5.0$ from the set $\{1.0, 2.0, \ldots, 6.0\}$.
However, we do not retune the number of hidden units on
\textsc{RoboCup}. For each iteration, we randomly sample a mini-batch
of $100$ scenarios during back-propagation and use
Adam~\cite{kingma-15} for optimization. Training typically converges
within $30$ epochs. We select the model according to the BLEU score on
the development set.\footnote{We implement our model in
  Theano~\cite{theano-10,theano-12} and will make the code publicly
  available.}

\paragraph{Evaluation Metrics} 
We consider two metrics as a means of evaluating the effectiveness of
our model on the two selective generation subproblems. For content
selection, we use the F-1 score of the set of selected records as
defined by the harmonic mean of precision and recall with respect to
the ground-truth selection record set. We define the set of selected
records as consisting of the record with the largest selection weight
$\alpha_{ti}$ computed by our aligner at each decoding step $t$.

We evaluate the quality of surface realization using the BLEU
score\footnote{We compute BLEU using the publicly available evaluation
  provided by \newcite{angeli-10}.}  (a $4$-gram matching-based
precision)~\cite{papineni-01} of the generated description with
respect to the human-created reference. To be comparable to previous
results on \textsc{WeatherGov}, we also consider a modified BLEU score
(cBLEU) that does not penalize numerical deviations of at most
five~\cite{angeli-10} (i.e., to not penalize ``low around 58''
compared to a reference ``low around 60''). On \textsc{RoboCup}, we
also evaluate the BLEU score in the case that ground-truth content
selection is known (sBLEU$_{\text{G}}$), to be comparable to previous
work.
\section{Results and Analysis}
\label{sec:results}

We analyze the effectiveness of our model on the benchmark
\textsc{WeatherGov} (as primary) and \textsc{RoboCup} (as
generalization) datasets. We also present several ablations to
illustrate the contributions of the primary model components.

\subsection{Primary Results (\textsc{WeatherGov})} \label{sec:primary}

We report the performance of content selection and surface realization
using F-1 and two BLEU scores (standard sBLEU and the customized cBLEU
of \newcite{angeli-10}), respectively
(\secref{sec:experimentalsetup}). Table~\ref{weatherresult} compares
our test results against previous methods that include
KL12~\cite{konstas-12}, KL13~\cite{konstas-13}, and
ALK10~\cite{angeli-10}.  Our method achieves the best results reported
to-date on all three metrics, with relative improvements of $11.94\%$
(F-1), $58.88\%$ (sBLEU), and $36.68\%$ (cBLEU) over the previous
state-of-the-art.
\begin{table}[!t]
    \centering
    {\small
    \caption{Primary \textsc{WeatherGov} results}\label{weatherresult}
    \begin{tabularx}{0.775\linewidth}{l c c c}
        \toprule
		Method & F-1 & sBLEU & cBLEU \\
        \midrule
		\hyperlinkcite{konstas-12}{KL12} & -- & $33.70$ & -- \\
        \hyperlinkcite{konstas-13}{KL13} & -- & $36.54$ & -- \\
        \hyperlinkcite{angeli-10}{ALK10} & $65.40$ & $38.40$ & $51.50$ \\
        Our model & $\mathbf{73.21}$ & $\mathbf{61.01}$ & $\mathbf{70.39}$ \\
        \bottomrule
    \end{tabularx}}
\end{table}

\subsection{Beam Filter with $k$-Nearest
  Neighbors} \label{sec:extensions}

We considered beam search as an alternative to greedy search in our
primary setup (Eqn.~\ref{eqn:argmax}), but this performs worse,
similar to what previous work found on this dataset~\cite{angeli-10}.
As an alternative, we consider a beam filter based on a $k$-nearest
neighborhood. See \hyperref[sec:appendix]{Supplementary Material} for details.
Table~\ref{knntest} shows that this $k$-NN beam filter improves
results over the primary greedy results.
\begin{table}[!ht]
    \centering
    {\small
    \caption{$k$-NN beam filter (test set)}\label{knntest}
    \begin{tabularx}{0.85\linewidth}{l c c }
        \toprule
		 & \ \ \ \  Primary &  \ \ \ \ $k$-NN Beam Filter \\
        \midrule
        sBLEU & $61.01$ & $\mathbf{61.76}$ \\
        cBLEU & $70.39$ & $\mathbf{71.23}$ \\
        \bottomrule
    \end{tabularx}}
\end{table}
\begin{figure*}
    \centering
    \includegraphics[width=1.0\textwidth]{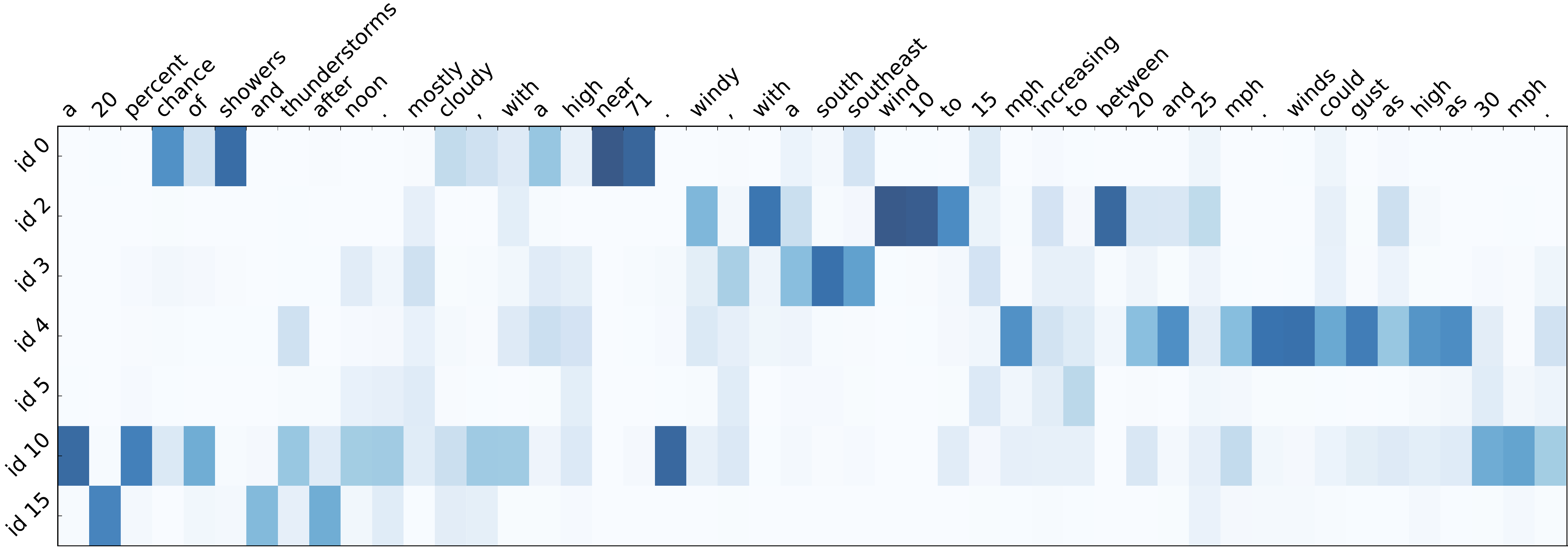}
    \begin{minipage}[!t]{1.0\linewidth}
        \begin{mdframed}[align=left]{\setlength{\baselineskip}%
              {0.8\baselineskip}{\scriptsize
                Record details:\\
                id-0:
                temperature(time=\texttt{06-21}, min=\texttt{52},
                mean=\texttt{63}, max=\texttt{71}); \quad
                id-2: windSpeed(time=\texttt{06-21}, min=\texttt{8}, 
                mean=\texttt{17}, max=\texttt{23});\\
                id-3: windDir(time=\texttt{06-21}, mode=\texttt{SSE});  \quad
                id-4: gust(time=\texttt{06-21},
                min=\texttt{0}, mean=\texttt{10}, max=\texttt{30});\\
                id-5: skyCover(time=\texttt{6-21},
                mode=\texttt{50-75});  \quad
                id-10: precipChance(time=\texttt{06-21},
                min=\texttt{19}, mean=\texttt{32}, max=\texttt{73});\\
                id-15: thunderChance(time=\texttt{13-21}, mode=\texttt{SChc})}\par}
        \end{mdframed}
    \end{minipage}
    {\small\caption{An example generation for a set of records from \textsc{WeatherGov}.}\label{fig:outputfig}\vspace{-0.25cm}}
\end{figure*}

\subsection{Ablation Analysis (\textsc{WeatherGov})} \label{sec:ablation}

Next, we present several ablations to analyze the contribution of our
model components.\footnote{These results are based on our primary
  model of Sec.~\ref{sec:primary} and on the development set.}

\paragraph{Aligner Ablation}
First, we evaluate the contribution of our proposed coarse-to-fine
aligner by comparing our model with the basic encoder-aligner-decoder
model introduced by \newcite{bahdanau-14}.  Table~\ref{devresults}
reports the results demonstrating that our aligner yields superior F-1
and BLEU scores relative to a standard aligner.
\begin{table}[!ht]
    \centering
    {\small
      \caption{Coarse-to-fine aligner ablation (dev
        set)}\label{devresults}
      \begin{tabularx}{0.825\linewidth}{l c c c}
        \toprule
        Method & F-1 & sBLEU & cBLEU \\
        \midrule
        Basic & $60.35$ & $63.54$ & $74.90$ \\
        Coarse-to-fine & $76.28$ & $65.58$ & $75.78$ \\
        \bottomrule
      \end{tabularx}}
\end{table}

\paragraph{Encoder Ablation} 
Next, we consider the effectiveness of the encoder.
Table~\ref{noencoder} compares the results with and without the
encoder on the development set, and demonstrates that there is a
significant gain from encoding the event records using the LSTM-RNN.
We attribute this improvement to the LSTM-RNN's ability to capture the
relationships that exist among the records, which is known to be
essential to selective generation~\cite{barzilay-05,angeli-10}.
\begin{table}[!ht]
    \centering
    {\small
    \caption{Encoder ablation (dev set)}\label{noencoder}
    \begin{tabularx}{0.725\linewidth}{l c c c}
      \toprule
      Encoder & F-1 & sBLEU & cBLEU \\
      \midrule
      With & $76.28$ & $65.58$ & $75.78$ \\
      Without & $57.45$ & $56.47$ & $68.63$ \\
      \bottomrule
    \end{tabularx}}
\end{table}

\subsection{Qualitative Analysis (\textsc{WeatherGov})}

\paragraph{Output Examples}
Fig.~\ref{fig:outputfig} shows an example record set with its output
description and record-word alignment heat map.  As shown, our model
learns to align records with their corresponding words (e.g., windDir
and ``southeast,'' temperature and ``71,'' windSpeed and ``wind 10,''
and gust and ``winds could gust as high as 30 mph'').  It also learns
the subset of salient records to talk about (matching the ground-truth
description perfectly for this example, i.e., a standard BLEU of
$100.00$).  We also see some word-level mismatch, e.g., ``cloudy''
mis-aligns to id-0 temp and id-10 precipChance, which we attribute to
the high correlation between these types of records (``garbage
collection'' in \newcite{liang-09}).

\paragraph{Word Embeddings}
Training our decoder has the effect of learning embeddings for the
words in the training set (via the embedding matrix $E$ in
\eqnref{eqn:decoder}). Here, we explore the extent to which these
learned embeddings capture semantic relationships among the training
words. Table~\ref{nntable} presents nearest neighbor words for some of
the common words from the \textsc{WeatherGov} dataset (according to
cosine similarity in the embedding space). More details of other
embedding approaches that we tried are discussed in the
\hyperref[sec:appendix]{Supplementary Material} section.
\begin{table}[!ht]
    \centering
    {\small
      \caption{Nearest neighbor word for example words}\label{nntable}
      \begin{tabularx}{0.59\linewidth}{c c}
        \toprule
        Word & Nearest neighbor \\
        \midrule
        gusts & gust \\
        clear & sunny \\
        isolated & scattered \\
        southeast & northeast  \\
        storms & winds  \\
        decreasing & falling \\
        \bottomrule
      \end{tabularx}}
\end{table}

\subsection{Out-of-Domain Results (\textsc{RoboCup})}

We use the \textsc{RoboCup} dataset to evaluate the
domain-independence of our model. The dataset is
severely data-starved with only $1000$ (approx.) training pairs, which
is much smaller than is typically necessary to train RNNs. This
results in higher variance in the trained model distributions, and we
thus adopt the
standard denoising method of
ensembles~\cite{sutskever-14,vinyals-14,zaremba-14}.\footnote{We use
  an ensemble of five randomly initialized models.}

\begin{table}[!ht]
    \centering
    {\small
    \caption{\textsc{RoboCup} results}\label{roboresult}
    \begin{tabularx}{0.8\linewidth}{l c c c}
        \toprule
		Method & F-1 & sBLEU & sBLEU$_{\text{G}}$\\
        \midrule
        \hyperlinkcite{chen-08}{CM08} & $72.00$  & -- & $28.70$\\
        \hyperlinkcite{liang-09}{LJK09} & $75.70$ & -- & -- \\
        \hyperlinkcite{chen-10}{CKM10} & $79.30$ & -- & --\\
        \hyperlinkcite{angeli-10}{ALK10} & $79.90$ & -- & $28.80$ \\
        \hyperlinkcite{konstas-12}{KL12} & --  & $24.88$ & $\mathbf{30.90}$\\
        Our model & $\mathbf{81.58}$ & $\mathbf{25.28}$ & $29.40$ \\
        \bottomrule
    \end{tabularx}}
\end{table}
Following previous work, we perform two experiments on the
\textsc{RoboCup} dataset (Table~\ref{roboresult}), the first
considering full selective generation and the second assuming
ground-truth content selection at test time. On the former, we obtain
a standard BLEU score (sBLEU) of $25.28$, which exceeds the best score
of $24.88$~\cite{konstas-12}. Additionally, we achieve an selection
F-1 score of $81.58$, which is also the best result reported
to-date. In the case of assumed (known) ground-truth content
selection, our model attains an sBLEU$_{\text{G}}$ score of $29.40$,
which is competitive with the state-of-the-art.\footnote{The
  \newcite{chen-08} sBLEU$_{\text{G}}$ result is from
  \newcite{angeli-10}.}
\section{Conclusion} \label{sec:conclusion}

We presented an encoder-aligner-decoder model for selective generation that does not use any specialized features, linguistic resources, or generation templates. Our model employs a bidirectional LSTM-RNN model with a novel coarse-to-fine aligner that jointly learns content selection and surface realization. We evaluate our model on the benchmark \textsc{WeatherGov} dataset and achieve state-of-the-art selection and generation results. We achieve further improvements via a $k$-nearest neighbor beam filter. We also present several model ablations and visualizations to elucidate the effects of the primary components of our model.  Moreover, our model generalizes to a different, data-starved domain (\textsc{RoboCup}), where it achieves results competitive with or better than the state-of-the-art.
\section*{Acknowledgments}

We thank Gabor Angeli, David Chen, and Ioannis Konstas for their helpful comments.


\appendix
\section{Supplementary Material} \label{sec:appendix}

The following provides further evaluations of our model as a
supplement to our original manuscript.

\subsection{Beam Filter with $k$-Nearest Neighbors} \label{sec:extensions-supp}

We perform greedy search as an approximation to full inference over the
set of decision variables (Eqn.~\ref{eqn:argmax}). We considered beam
search as an alternative, but as with previous work on this
dataset~\cite{angeli-10}, we found that greedy search still yields better
BLEU performance (Table~\ref{beamsearch}).
\begin{table}[!ht]
    \centering
    {\small
    \caption{Effect of beam width}\label{beamsearch}
    \begin{tabularx}{0.94\linewidth}{l c c c c }
        \toprule
		Beam width $M$ & $1$ & $2$ & $5$ & $10$ \\
        \midrule
        dev sBLEU & $65.58$ & $64.70$ & $57.02$ & $47.07$ \\
        dev cBLEU & $75.78$ & $74.91$ & $65.83$ & $54.19$ \\
        test sBLEU & $61.01$ & $60.15$ & $53.70$ & $44.32$ \\
        test cBLEU & $70.39$ & $69.42$ & $61.95$ & $51.01$ \\
        \bottomrule
    \end{tabularx}}
\end{table}

As an alternative, we consider a beam filter based on a $k$-nearest
neighborhood. First, we generate the $M$-best description candidates
(i.e., a beam width of $M$) for a given input record set (database) using
standard beam search. Next, we find the $K$ nearest neighbor
database-description pairs from the training data, based on the cosine
similarity of each neighbor database with the given input record.  We
then compute the BLEU score for each of the $M$ description candidates
relative to the $K$ nearest neighbor descriptions (as
references) and select the candidate with the highest BLEU score.
We tune $K$ and $M$ on the development set and report the results
in Table~\ref{knndev}. Table~\ref{knntest} presents the test
results with this tuned setting ($M=2$, $K=1$), where we achieve BLEU
scores better than our primary greedy results.
\begin{table}[!th]
    \centering
    {\small
    \caption{$k$-NN beam filter (dev set)}\label{knndev}
    \begin{tabularx}{0.75\linewidth}{l c c c }
        \toprule
		sBLEU & $M = 2$ & $M = 5$ &$ M = 10$ \\
        \midrule
        $K = 1$ & $\mathbf{65.99}$ & $65.88$ & $65.65$ \\
        $K = 2$ & $65.89$ & $65.98$ & $65.83$ \\
        $K = 5$ & $65.64$ & $65.45$ & $65.41$ \\
        $K = 10$ & $65.91$ & $65.89$ & $65.12$ \\
        \bottomrule
        \toprule
		cBLEU & $M = 2$ & $M = 5$ &$ M = 10$ \\
        \midrule
        $K = 1$ & $\mathbf{76.21}$ & $76.13$ & $75.98$ \\
        $K = 2$ & $75.99$ & $76.03$ & $75.82$ \\
        $K = 5$ & $75.90$ & $75.63$ & $75.41$ \\
        $K = 10$ & $75.95$ & $75.87$ & $75.23$ \\
        \bottomrule
    \end{tabularx}}
\end{table}
\begin{table}[!ht]
    \centering
    {\small
    \caption{$k$-NN beam filter (test set)}\label{knntest}
    \begin{tabularx}{0.85\linewidth}{l c c }
        \toprule
		 & Primary & $k$-NN ($M=2$, $K=1$) \\
        \midrule
        sBLEU & $61.01$ & $\mathbf{61.76}$ \\
        cBLEU & $70.39$ & $\mathbf{71.23}$ \\
        \bottomrule
    \end{tabularx}}
\end{table}

\subsection{Word Embeddings (Trained \& Pretrained)}
Training our decoder has the effect of learning embeddings for the
words in the training set (via the embedding matrix $E$ in
Eqn.~\ref{eqn:decoder}). Here, we explore the extent to which these
learned embeddings capture semantic relationships among the training
words. Table~\ref{nntable} presents nearest neighbor words for some of
the common words from the \textsc{WeatherGov} dataset (according
to cosine similarity in the embedding space).
\begin{table}[!ht]
\centering
{\small
    \caption{Nearest neighbor word for example words}\label{nntable}
    \begin{tabularx}{0.59\linewidth}{c c}
      \toprule
      Word & Nearest neighbor \\
      \midrule
      gusts & gust \\
      clear & sunny \\
      isolated & scattered \\
      southeast & northeast  \\
      storms & winds  \\
      decreasing & falling \\
      \bottomrule
    \end{tabularx}}
\end{table}

We also consider different ways of using pre-trained word
embeddings~\cite{mikolov-13} to bootstrap the quality of our learned
embeddings. One approach initializes our embedding matrix with the
pre-trained vectors and then refines the embedding based on
our training corpus. The second concatenates our learned embedding
matrix with the pre-trained vectors in an effort to simultaneously
exploit general similarities as well as those learned for the
domain. As shown previously for other
tasks~\cite{vinyals-14b,vinyals-14}, we find that the use of 
pre-trained embeddings results in negligible improvements (on the
development set).

\bibliographystyle{naaclhlt2016}
\bibliography{references}

\begin{thebibliography}{}

\bibitem[\protect\citename{Angeli \bgroup et al.\egroup }2010]{angeli-10}
Gabor Angeli, Percy Liang, and Dan Klein.
\newblock 2010.
\newblock A simple domain-independent probabilistic approach to generation.
\newblock In {\em Proceedings of the Conference on Empirical Methods in Natural
  Language Processing (EMNLP)}, pages 502--512.

\bibitem[\protect\citename{Bahdanau \bgroup et al.\egroup }2014]{bahdanau-14}
Dzmitry Bahdanau, Kyunghyun Cho, and Yoshua Bengio.
\newblock 2014.
\newblock Neural machine translation by jointly learning to align and
  translate.
\newblock {\em arXiv preprint arXiv:1409.0473}.

\bibitem[\protect\citename{Barzilay and Lapata}2005]{barzilay-05}
Regina Barzilay and Mirella Lapata.
\newblock 2005.
\newblock Collective content selection for concept-to-text generation.
\newblock In {\em Proceedings of the Human Language Technology Conference and
  the Conference on Empirical Methods in Natural Language Processing
  (HLT/EMNLP)}, pages 331--338.

\bibitem[\protect\citename{Barzilay and Lee}2004]{barzilay-04}
Regina Barzilay and Lillian Lee.
\newblock 2004.
\newblock Catching the drift: {P}robabilistic content models, with applications
  to generation and summarization.
\newblock In {\em Proceedings of the Conference of the North American Chapter
  of the Association for Computational Linguistics – Human Language
  Technologies (NAACL HLT)}, pages 113--120.

\bibitem[\protect\citename{Bastien \bgroup et al.\egroup }2012]{theano-12}
Fr{\'{e}}d{\'{e}}ric Bastien, Pascal Lamblin, Razvan Pascanu, James Bergstra,
  Ian~J. Goodfellow, Arnaud Bergeron, Nicolas Bouchard, and Yoshua Bengio.
\newblock 2012.
\newblock Theano: new features and speed improvements.
\newblock NIPS Workshop on Deep Learning and Unsupervised Feature Learning.

\bibitem[\protect\citename{Belz}2008]{belz-08}
Anja Belz.
\newblock 2008.
\newblock Automatic generation of weather forecast texts using comprehensive
  probabilistic generation-space models.
\newblock {\em Natural Language Engineering}, 14(04):431--455.

\bibitem[\protect\citename{Bergstra \bgroup et al.\egroup }2010]{theano-10}
James Bergstra, Olivier Breuleux, Fr{\'{e}}d{\'{e}}ric Bastien, Pascal Lamblin,
  Razvan Pascanu, Guillaume Desjardins, Joseph Turian, David Warde-Farley, and
  Yoshua Bengio.
\newblock 2010.
\newblock Theano: a {CPU} and {GPU} math expression compiler.
\newblock In {\em Proceedings of the Scientific Computing with Python
  Conference (SciPy)}.

\bibitem[\protect\citename{Chen and Mooney}2008]{chen-08}
David~L. Chen and Raymond~J. Mooney.
\newblock 2008.
\newblock Learning to sportscast: a test of grounded language acquisition.
\newblock In {\em Proceedings of the International Conference on Machine
  Learning (ICML)}, pages 128--135.

\bibitem[\protect\citename{Chen \bgroup et al.\egroup }2010]{chen-10}
David~L. Chen, Joohyun Kim, and Raymond~J. Mooney.
\newblock 2010.
\newblock Training a multilingual sportscaster: Using perceptual context to
  learn language.
\newblock {\em Journal of Artificial Intelligence Research}, 37:397--435.

\bibitem[\protect\citename{Graves \bgroup et al.\egroup }2013]{graves-13}
Alex Graves, Mohamed Abdel-rahman, and Geoffrey Hinton.
\newblock 2013.
\newblock Speech recognition with deep recurrent neural networks.
\newblock In {\em Proceedings of the IEEE International Conference on
  Acoustics, Speech and Signal Processing (ICASSP)}, pages 6645--6649.

\bibitem[\protect\citename{Hochreiter and Schmidhuber}1997]{hochreiter-97}
Sepp Hochreiter and J{\"u}rgen Schmidhuber.
\newblock 1997.
\newblock Long short-term memory.
\newblock {\em Neural Computation}, 9(8):1735--1780.

\bibitem[\protect\citename{Karpathy and Fei-Fei}2015]{karpathy-15}
Andrej Karpathy and Li~Fei-Fei.
\newblock 2015.
\newblock Deep visual-semantic alignments for generating image descriptions.
\newblock In {\em Proceedings of the IEEE Conference on Computer Vision and
  Pattern Recognition (CVPR)}, pages 3128--3137.

\bibitem[\protect\citename{Kim and Mooney}2010]{kim-10}
Joohyun Kim and Raymond~J Mooney.
\newblock 2010.
\newblock Generative alignment and semantic parsing for learning from ambiguous
  supervision.
\newblock In {\em Proceedings of the International Conference on Computational
  Linguistics (COLING)}, pages 543--551.

\bibitem[\protect\citename{Kingma and Ba}2015]{kingma-15}
Diederik Kingma and Jimmy Ba.
\newblock 2015.
\newblock Adam: A method for stochastic optimization.
\newblock In {\em Proceedings of the International Conference on Learning
  Representations (ICLR)}.

\bibitem[\protect\citename{Konstas and Lapata}2012]{konstas-12}
Ioannis Konstas and Mirella Lapata.
\newblock 2012.
\newblock Unsupervised concept-to-text generation with hypergraphs.
\newblock In {\em Proceedings of the Conference of the North American Chapter
  of the Association for Computational Linguistics – Human Language
  Technologies (NAACL HLT)}, pages 752--761.

\bibitem[\protect\citename{Konstas and Lapata}2013]{konstas-13}
Ioannis Konstas and Mirella Lapata.
\newblock 2013.
\newblock Inducing document plans for concept-to-text generation.
\newblock In {\em Proceedings of the Conference on Empirical Methods in Natural
  Language Processing (EMNLP)}, pages 1503--1514.

\bibitem[\protect\citename{Liang \bgroup et al.\egroup }2009]{liang-09}
Percy Liang, Michael~I. Jordan, and Dan Klein.
\newblock 2009.
\newblock Learning semantic correspondences with less supervision.
\newblock In {\em Proceedings of the Joint Conference of the Annual Meeting of
  the Association for Computational Linguistics and the International Joint
  Conference on Natural Language Processing (ACL/IJCNLP)}, pages 91--99.

\bibitem[\protect\citename{Lu and Ng}2011]{lu-11}
Wei Lu and Hwee~Tou Ng.
\newblock 2011.
\newblock A probabilistic forest-to-string model for language generation from
  typed lambda calculus expressions.
\newblock In {\em Proceedings of the Conference on Empirical Methods in Natural
  Language Processing (EMNLP)}, pages 1611--1622.

\bibitem[\protect\citename{Lu \bgroup et al.\egroup }2008]{lu-08}
Wei Lu, Hwee~Tou Ng, Wee~Sun Lee, and Luke~S Zettlemoyer.
\newblock 2008.
\newblock A generative model for parsing natural language to meaning
  representations.
\newblock In {\em Proceedings of the Conference on Empirical Methods in Natural
  Language Processing (EMNLP)}, pages 783--792.

\bibitem[\protect\citename{Lu \bgroup et al.\egroup }2009]{lu-09}
Wei Lu, Hwee~Tou Ng, and Wee~Sun Lee.
\newblock 2009.
\newblock Natural language generation with tree conditional random fields.
\newblock In {\em Proceedings of the Conference on Empirical Methods in Natural
  Language Processing (EMNLP)}, pages 400--409.

\bibitem[\protect\citename{Mei \bgroup et al.\egroup }2015]{mei-15}
Hongyuan Mei, Mohit Bansal, and Matthew~R. Walter.
\newblock 2015.
\newblock Listen, attend, and walk: {N}eural mapping of navigational
  instructions to action sequences.
\newblock {\em arXiv preprint arXiv:1506.04089}.

\bibitem[\protect\citename{Mikolov \bgroup et al.\egroup }2013]{mikolov-13}
Tomas Mikolov, Kai Chen, Greg Corrado, and Jeffrey Dean.
\newblock 2013.
\newblock Efficient estimation of word representations in vector space.
\newblock In {\em Proceedings of the International Conference on Learning
  Representations (ICLR)}.

\bibitem[\protect\citename{Papineni \bgroup et al.\egroup }2001]{papineni-01}
Kishore Papineni, Salim Roukos, Todd Ward, and Wei-Jing Zhu.
\newblock 2001.
\newblock {BLEU}: a method for automatic evaluation of machine translation.
\newblock In {\em Proceedings of the Annual Meeting of the Association for
  Computational Linguistics (ACL)}, pages 311--318.

\bibitem[\protect\citename{Pascanu \bgroup et al.\egroup }2014]{pascanu-14}
Razvan Pascanu, Caglar Gulcehre, Kyunghyun Cho, and Yoshua Bengio.
\newblock 2014.
\newblock How to construct deep recurrent neural networks.
\newblock In {\em Proceedings of the International Conference on Learning
  Representations (ICLR)}.

\bibitem[\protect\citename{Soricut and Marcu}2006]{soricut-06}
Radu Soricut and Daniel Marcu.
\newblock 2006.
\newblock Stochastic language generation using {WIDL}-expressions and its
  application in machine translation and summarization.
\newblock In {\em Proceedings of the International Conference on Computational
  Linguistics and the Annual Meeting of the Association for Computational
  Linguistics (COLING/ACL)}, pages 1105--1112.

\bibitem[\protect\citename{Sutskever \bgroup et al.\egroup }2014]{sutskever-14}
Ilya Sutskever, Oriol Vinyals, and Quoc~V. Lee.
\newblock 2014.
\newblock Sequence to sequence learning with neural networks.
\newblock In {\em Advances in Neural Information Processing Systems (NIPS)}.

\bibitem[\protect\citename{Vinyals \bgroup et al.\egroup }2014]{vinyals-14b}
Oriol Vinyals, Lukasz Kaiser, Terry Koo, Slav Petrov, Ilya Sutskever, and
  Geoffrey Hinton.
\newblock 2014.
\newblock Grammar as a foreign language.
\newblock {\em arXiv preprint arXiv:1412.7449}.

\bibitem[\protect\citename{Vinyals \bgroup et al.\egroup }2015a]{vinyals-15}
Oriol Vinyals, Samy Bengio, and Manjunath Kudlur.
\newblock 2015a.
\newblock Order matters: {S}equence to sequence for sets.
\newblock {\em arXiv preprint arXiv:1511.06391}.

\bibitem[\protect\citename{Vinyals \bgroup et al.\egroup }2015b]{vinyals-14}
Oriol Vinyals, Alexander Toshev, Samy Bengio, and Dumitru Erhan.
\newblock 2015b.
\newblock Show and tell: {A} neural image caption generator.
\newblock In {\em Proceedings of the IEEE Conference on Computer Vision and
  Pattern Recognition (CVPR)}, pages 3156--3164.

\bibitem[\protect\citename{Wong and Mooney}2007]{wong-07}
Yuk~Wah Wong and Raymond~J Mooney.
\newblock 2007.
\newblock Generation by inverting a semantic parser that uses statistical
  machine translation.
\newblock In {\em Proceedings of the Conference of the North American Chapter
  of the Association for Computational Linguistics – Human Language
  Technologies (NAACL HLT)}, pages 172--179.

\bibitem[\protect\citename{Xu \bgroup et al.\egroup }2015]{xu-15}
Kelvin Xu, Jimmy Ba, Ryan Kiros, Kyunghyun Cho, Aaron Courville, Ruslan
  Salakhutdinov, Richard Zemel, and Yoshua Bengio.
\newblock 2015.
\newblock Show, attend and tell: Neural image caption generation with visual
  attention.
\newblock In {\em Proceedings of the International Conference on Machine
  Learning (ICML)}.

\bibitem[\protect\citename{Zaremba \bgroup et al.\egroup }2014]{zaremba-14}
Wojciech Zaremba, Ilya Sutskever, and Oriol Vinyals.
\newblock 2014.
\newblock Recurrent neural network regularization.
\newblock {\em arXiv preprint arXiv:1409.2329}.

\end{thebibliography}

\end{document}